# Benchmarking AI for low-resource contexts: Thinking beyond leaderboards

A system-level white paper on how to benchmark speech, chat, and vision systems for real deployment conditions in low-resource settings.

Aakash Pant*, Kavya Shah*, Apoorv Agnihotri*, Sneha Nikam, Prasaanth Balraj, Nakul Jain

*Wadhwani AI Global*

**Abstract**

Existing AI evaluation practices often fail to capture how systems actually perform in low-resource environments, where operational constraints shape usability as much as model quality. Through a structured analysis of existing benchmark families across speech, chat/RAG, and vision systems, we identify critical gaps between laboratory evaluation practices and real-world deployment conditions in low-resource environments. We argue that the meaningful unit of assessment is the deployed system rather than an isolated model and that effective evaluation frameworks must integrate task performance with deployment conditions such as noisy inputs, code-switching, intermittent connectivity, low-end hardware, and domain shift. At the same time, benchmarks should recognize that different application classes require distinct evaluation profiles rather than a single aggregate score that obscures operational differences. To support practical decision-making, we propose a shared reporting framework that preserves comparability across systems and application types while remaining sensitive to deployment context. Finally, we emphasize the need for concise and actionable reporting artifacts for policymakers, donors, and implementers, including standardized one-page benchmark cards, deployment profiles, and explicit documentation of failure handling procedures and human oversight mechanisms.

## 1. Reframing the problem

Most AI leaderboards were designed to compare models under standardized laboratory conditions. While useful for research progress, such comparisons are not enough to guide deployment decisions in low-resource settings.

In practice, users do not interact with a model in isolation. They interact with a system

**These authors contributed equally to this work.*

in which several factors, including data capture, preprocessing, retrieval, orchestration, inference, user interface, and fallback logic, shape the final outcome. This gap between a model's performance under benchmarking conditions and real-world deployment matters more in low-resource environments because device quality, network reliability, and input capture conditions vary more widely and are more likely to influence outcomes. The International Telecommunication Union (ITU) estimates that about 2.2 billion people remained offline in 2025, most of them in low- and middle-income countries, and World Bank analysis highlights device affordability as a major barrier even where mobile broadband coverage exists. [1], [2]

Under these conditions, a benchmark that ignores device class, network quality, and use context can misstate the deployment readiness of a model and its surrounding system. For example, speech systems that perform well under controlled benchmarking conditions may degrade substantially in real-world deployments due to code-switching, far-field audio capture, low-quality microphones, or intermittent connectivity. Similarly, image-based systems evaluated on curated datasets may underperform when exposed to low-light conditions, motion blur, or device variability commonly observed in frontline deployment environments.

In deployments of AI-supported chronic pulmonary obstructive disease (COPD) screening systems in India, speech and cough-based models encountered operational conditions that were poorly represented in laboratory benchmarks, including background noise, variable microphone quality, multilingual interactions, and inconsistent recording environments across field sites. While baseline model metrics remained stable under controlled evaluation settings, workflow usability and prediction reliability were more sensitive to real-world capture conditions, operator variability, and deployment infrastructure constraints. These realities reinforce the need for deployment-oriented benchmarking frameworks that evaluate complete operational systems rather than isolated models. [3], [4]

This paper draws on similar observations from AI field deployments across public health settings in India to argue that **foundational AI should be benchmarked as end-to-end systems, not just as standalone models.** [4] This perspective does not replace model-level benchmarking; instead, it situates model evaluation within a broader systems framework that reflects how AI is actually experienced in real-world operational settings.

## 2. What the current benchmark landscape offers

The existing benchmark ecosystem is valuable, but fragmented by design. Different benchmark families address different parts of AI systems, and each applies to different AI components including large language models (LLMs), automatic speech recognition (ASR) systems, chat pipelines, speech translation systems, and vision models.

### From Benchmarks to Field-Readiness: An Application Crosswalk

The matrix below organizes benchmark families by application type, highlighting both the contributions of each family and the deployment tests still required to assess a given system for field-readiness.

**Table 1. Cataloguing benchmark requirements by application type**

| Application type | Recommended benchmark families | What they evaluate | Required system tests beyond the benchmark (non-exhaustive) |
|---|---|---|---|
| Automatic speech recognition (ASR) / speech translation | • FLEURS [5]<br>• Speech Robust Bench [6]<br>• Local multilingual and code-switching test sets [7] | **FLEURS**: multilingual ASR, speech language identification, speech translation, and speech/text retrieval.<br><br>**Speech Robust Bench:** ASR robustness under acoustic and signal corruptions.<br><br>**Local sets**: target-domain language coverage, accents, dialects, and code-switching. | • Noise bands<br>• Microphone variability<br>• On-device latency<br>• Offline fallback<br>• Accent and dialect transfer<br>• Connectivity drop tests |
| Chat, QA, and search systems | • HELM or HELM Instruct [8], [9]<br>• RAGAs [10]<br>• CRAG [11], [12] | **HELM / HELM Instruct:** general LLM behavior, instruction following, calibration, | • Low-bandwidth behavior<br>• Timeout and retry handling |

| Application type | Recommended benchmark families | What they evaluate | Required system tests beyond the benchmark (non-exhaustive) |
|---|---|---|---|
| | ● Domain-grounded evaluation sets | robustness, and performance across prompting and task conditions.<br><br>**RAGAs:** evaluation of RAG pipelines, including answer relevance, faithfulness, context precision/recall, grounding, and robustness to noisy retrieval.<br><br>**CRAG:** factual question answering under realistic retrieval conditions, including dynamic knowledge, ambiguous retrieval, and noisy or incomplete retrieved context.<br><br>**Domain-grounded sets:** target-domain terminology, workflow fidelity, localized knowledge coverage, and site-specific reasoning performance. | ● Cache behavior<br>● Offline fallback where applicable<br>● Human escalation rules<br>● Domain-shift transfer across sites<br>● Retrieval failure and empty-context handling<br>● Citation and source-trace reliability |
| Vision classification / screening | ● ImageNet-C/ImageNet-P-style corruption and perturbation testing [13] | **ImageNet-C / ImageNet-P–style testing:** robustness to common image corruptions and | ● Capture-quality checks<br>● Low-light and blur stress<br>● Device camera |

| Application type | Recommended benchmark families | What they evaluate | Required system tests beyond the benchmark (non-exhaustive) |
| --- | --- | --- | --- |
| | ● Local clinical or field-image holdouts<br>● Subgroup and site-stratified validation sets | perturbations, including noise, blur, compression artifacts, and distribution shift.<br><br>**Local holdouts:** target-domain image characteristics, device variability, environmental conditions, and out-of-domain generalization across deployment settings.<br><br>**Subgroup and site-stratified validation:** performance variation across demographic groups, acquisition sites, devices, and operational contexts.<br><br>Task-specific evaluation of AUROC, AUPRC, sensitivity/specificity, PPV/NPV, F1, and calibration. | variation<br>● Referral thresholds<br>● Unusable-image detection<br>● Workflow fallback when images are poor quality or outside the intended use case |
| Vision segmentation / detection | ● Task-specific segmentation and detection benchmarks<br>● ImageNet-inspired corruption testing, including | **Segmentation and detection benchmarks:** segmentation quality, object localization, boundary accuracy, and detection performance | ● Low-light and motion-blur stress testing<br>● Occlusion and partial-object robustness<br>● Small-object and |

| Application type | Recommended benchmark families | What they evaluate | Required system tests beyond the benchmark (non-exhaustive) |
| --- | --- | --- | --- |
| | task-specific variants such as COCO-C, PASCAL-C, Cityscapes-C, or segmentation corruption benchmarks [13], [14]<br>• Multi-site holdout sets | across task-specific settings.<br><br>**Corruption and perturbation testing:** robustness to noise, blur, compression artifacts, occlusion, and other image degradations affecting segmentation and detection performance.<br><br>**Multi-site holdouts:** generalization across acquisition sites, devices, imaging conditions, and deployment domains.<br><br>Task-specific performance evaluation using Dice coefficient, Intersection over Union (IoU), mean Average Precision (mAP), and Hausdorff or surface-distance metrics. | edge-case detection<br>• Unusable-image and artifact detection<br>• Workflow fallback for low-confidence or out-of-scope predictions<br>• Human review and escalation pathways |

The following sections address each application type separately, outlining an end-to-end benchmarking framework that accounts for deployment conditions.

## 3. The right unit of analysis: the system under test

While existing benchmark families provide valuable starting points, many remain limited in their representation of low-resource deployment realities, including multilingual code-switching behavior, dialect variation, low-end device environments, and geographically diverse operational conditions.

In several large-scale public health AI deployments, operational failure modes emerged outside traditional benchmark scope. For instance, health-related risk prediction systems may maintain strong predictive performance during retrospective evaluation, while degrading operationally in predictive settings due to incomplete field data synchronization, delayed reporting pipelines, or shifts in local healthcare-seeking behavior. Such deployment conditions highlight the importance of evaluating workflow resilience and operational robustness in addition to model accuracy. [15]

This system-first framing is aligned with the National Institute of Standards and Technology's (NIST) Artificial Intelligence Risk Management Framework (RMF), which treats AI as part of a socio-technical system and emphasizes mapping, measuring, and managing risks in deployment context. The framework presented below builds on the RMF by offering a practical mechanism by which to evaluate AI systems holistically for low-resource contexts. [16]

Benchmarking AI systems in such environments should begin by defining the system under test. That definition should include the user-facing workflow, along with every technical dependency that materially changes outcome quality. For a voice system, that usually means audio capture, language identification, speech recognition, downstream task logic, and any offline fallback. For a chat system, it includes retrieval, reranking, generation, citation logic, guardrails, and network dependencies. For a vision screening system, it includes image capture guidance, image-quality gates, model inference, thresholding, and referral logic.

Once the system boundary is fixed, benchmarks should cover the following three layers separately, rather than collapsing them into one score. This paper later introduces a model card designed to capture these layers.

**Table 2. Evaluating the components of a deployed AI system**

| Layer | Question answered | Typical metrics | Why it matters |
|---|---|---|---|
| Component layer | How well does each technical module work? | • Task-specific performance metrics (e.g., | Useful for isolating failures, comparing architectures, and |

| Layer | Question answered | Typical metrics | Why it matters |
|---|---|---|---|
| | | Word Error Rate (WER), retrieval recall, AUROC)<br>• Memory requirements<br>• Model size<br>• Calibration error<br>• Latency | optimizing technical performance. |
| Workflow layer | Does the full user task succeed? | • Task completion rate<br>• Answer correctness<br>• Referral correctness<br>• Ask-repeat rate<br>• Abstention rate<br>• Time to completion | Useful for evaluating whether the system actually supports successful task completion through the tested workflow. |
| Operating-condition layer | How much does performance degrade when the context gets hard? | • Performance delta under noise (e.g., missing data, code-switching, variable connectivity, device downgrade, site shift) | Useful for understanding deployment resilience under real-world constraints. |

# 4. Benchmark profiles by application type

A common benchmark standard should then use application-specific profiles to present evidence from speech, chat/RAG, and vision systems in a consistent structure across the component, workflow, and operating-condition layers.

## 4.1 Speech and voice systems

For speech systems, benchmarks should cover both recognition quality and service usability. FLEURS provides broad multilingual coverage for speech tasks, while Speech Robust Bench focuses explicitly on corruption robustness for ASR. [5], [6] In low-resource deployments, the required stress conditions should additionally include background noise, far-field capture, overlapping speech, microphone variation, accented speech, and expected code-switching.

| | |
|---|---|
| **System boundary** | Audio capture → Preprocessing → ASR and language handling → Downstream task logic → User feedback or retry path |
| **Mandatory stress tests** | • Noise bands<br>• Multiple microphones or phones<br>• Speaker diversity<br>• Code-switching<br>• Packet loss if cloud inference is used |
| **Core metrics** | • WER or Character Error Rate (CER)<br>• Task completion rate<br>• Ask-repeat rate<br>• Abstention or handoff rate<br>• 95th-percentile latency on the target device class |
| **What to report publicly** | • Reference-condition score<br>• Stressed-condition score<br>• Worst subgroup result<br>• Minimum tested device<br>• Network profile |

## 4.2 Chat and search systems

For chat and search systems, testing should cover both retrieval performance and downstream answer behavior. RAGAs was designed to evaluate retrieval-augmented pipelines without depending on human gold labels for every example. [10] CRAG adds more realistic factual question-answering with domain diversity and temporal variation. [11], [12] In low-resource settings, the benchmark must also simulate intermittent connectivity, stale local indexes, multilingual queries, and long-tail or time-sensitive queries because those conditions materially affect retrieval completeness and answer quality but are underrepresented in general leaderboards.

| | |
|---|---|
| **System boundary** | Retriever → Reranker → LLM → Citation or grounding layer → Safety rules → User interface and connectivity behavior |
| **Mandatory stress tests** | • Low bandwidth<br>• Request timeouts<br>• Stale or partially synced knowledge bases<br>• Multilingual prompts<br>• Long-tail or time-sensitive queries |
| **Core metrics** | • Answer correctness<br>• Faithfulness to source<br>• Citation precision<br>• Abstention rate<br>• Degraded-network success rate<br>• 95th-percentile latency |
| **What to report publicly** | • System behavior under missing evidence or retrieval failure<br>• Whether the system abstains or fabricates responses<br>• User-facing failure and fallback behavior |

### 4.3 Vision screening and image-based triage systems

For image-based screening systems, testing should cover real-world variability in image quality, capture conditions, and deployment environments. Corruption-style evaluation borrowed from ImageNet-C is useful for testing blur, noise, compression, and other capture degradations, while the WILDS benchmark datasets demonstrate why benchmark design should incorporate external validation across sites and domains from the outset. [13], [17] For high-stakes deployments, reporting should also align with domain standards such as TRIPOD+AI for prognostic prediction studies, STARD-AI for diagnostic accuracy studies, CLAIM for medical imaging studies, and CONSORT-AI for clinical trials involving AI. [18], [19], [20], [21] Though not benchmark suites or leaderboards, these reporting standards make evaluation evidence, external validation, and failure boundaries clearer in high-stakes settings.

| | |
|---|---|
| **System boundary** | Capture guidance → Image-quality gate → Model inference → Thresholding or triage rule → Referral or repeat-capture workflow |
| **Mandatory stress tests** | • Blur, glare, and low-light conditions<br>• Compression and image degradation artifacts |
| | • Device and phone-camera variation<br>• Prevalence and class-distribution shift<br>• Cross-site and domain shift<br>• Subgroup and demographic variation |
| **Core metrics** | • Sensitivity and specificity<br>• AUROC and task-specific operating-point metrics<br>• Calibration<br>• Rejection or abstention rate<br>• Subgroup- and site-stratified performance |
| **What to report publicly** | • Capture failure rate<br>• False-negative risk at the selected operating threshold<br>• External-site performance<br>• Human follow-up and escalation rules |

## 5. A minimum benchmark standard for low-resource deployment

The following reporting requirements form the practical core of the proposed standard for low-resource deployment.

1. **State the deployment profile.** Report the intended languages, geographies, device classes, network assumptions, and operator roles.
2. **Declare the system boundary.** Specify the components evaluated together, including third-party models, retrieval systems, client logic, and fallback behavior.
3. **Report operating conditions separately.** Publish results for reference, stressed, and severe operating conditions.
4. **Report both component-level and workflow-level results.** Include module metrics alongside end-to-end task and workflow performance.
5. **Provide shift and subgroup evaluation evidence.** Report performance across languages, sites, devices, and other relevant deployment subgroups, including out-of-domain testing where available. Include deployment-condition evidence from pilot or field implementations where available, particularly for workflows expected to operate under heterogeneous device, connectivity, or operator conditions
6. **Document failure handling and oversight behavior.** Report abstention logic,

retry behavior, human escalation pathways, and system deactivation triggers.

7. **Publish a benchmark card.** Accompany external claims with a benchmark card and supporting evaluation documentation, consistent with the spirit of model cards and datasheets. [22], [23] The benchmark card below (Table 3) operationalizes the reporting requirements described above in a concise, standardized format. It is designed for procurement committees, donors, and policy reviewers who need to compare systems without relying on a full technical appendix.

**Table 3. Proposed model card for AI systems deployed in low-resource contexts**

| Field | What should be reported |
|---|---|
| System name and version | Exact version tested, including major third-party components. |
| Intended use | What decision or workflow the system supports, and what it should not be used for. |
| Application profile | Speech / chat / vision screening / other. |
| Deployment context | Country or region, frontline user, device class, and connectivity assumptions. |
| System boundary | Which modules were included in the benchmark. |
| Reference-condition result | Primary module and workflow-level metric(s) under clean or nominal conditions. |
| Stress-condition results | Performance of the same workflow under disclosed operating-condition tests, such as noise, code-switching, device downgrade, weak network, or domain shift. |
| Worst observed subgroup or site result | Lowest disclosed subgroup or site result, reported separately from average stress-condition degradation. |
| Failure handling | Abstention, escalation, repeat-capture, or human-review rule. |
| Evidence status | Lab only / pilot validated / multisite validated. |
| Rollout recommendation | Not ready / pilot only / constrained rollout / scale candidate. |

## 6. Conclusion

In low-resource settings, a strong model inside a brittle system is still a brittle service. Thus, the phrase "beyond leaderboards" should not be read as anti-benchmark, but rather as a recognition that benchmarks should reflect the deployed unit.

Experience from real-world deployments across public health and frontline service delivery in India suggests that deployment constraints often emerge at the workflow and operational-condition layers rather than solely at the model/component layer. Benchmarking approaches that fail to capture these realities risk overstating a model's deployment readiness in low-resource settings. [4], [15]

As such, a practical standard should do three things at once: keep the discipline of model evaluation, add workflow-level evidence, and make operating conditions visible. When that structure is applied consistently across speech, chat/RAG, and vision systems, benchmark claims become more useful to engineering teams and more legible to policy and donor audiences.

The benchmark then becomes more relevant, revealing not which model looks best in the lab, but which system continues to perform as intended when field conditions are difficult.

## 7. Limitations and Future Works

This paper presents a deployment-oriented benchmarking framework and reporting structure informed by observations from real-world deployments. Future work should focus on empirically validating the proposed framework by assessing its utility across operational pilots involving speech, chat/RAG, and vision systems in diverse low-resource settings. Further research is also needed to expand benchmarking methods for multilingual and code-switching scenarios, where existing evaluation resources remain limited.